\documentclass[letterpaper]{article} 
\usepackage[preprint]{aaai2027}  

\usepackage{graphicx}
\usepackage{booktabs}
\usepackage{pifont}
\usepackage{amsthm,amsmath,amssymb}
\usepackage{multirow} 
\usepackage{enumitem}
\usepackage{booktabs} 
\usepackage{fontawesome}
\usepackage{array}
\usepackage{colortbl} 
\usepackage{xcolor} 
\usepackage{subcaption}
\newcommand{\up}[1]{\textsubscript{\color{green!70!black}{↑#1}}}
\newcommand{\down}[1]{\textsubscript{\color{red!70!black}{↓#1}}}

\usepackage{times}  
\usepackage{helvet}  
\usepackage{courier}  
\usepackage[hyphens]{url}  
\usepackage{graphicx} 
\usepackage{natbib}  
\usepackage{caption} 
\usepackage{algorithm}
\usepackage{algorithmic}

\usepackage{newfloat}
\usepackage{listings}
\DeclareCaptionStyle{ruled}{labelfont=normalfont,labelsep=colon,strut=off} 
\floatstyle{ruled}
\usepackage{enumitem}
\usepackage[most]{tcolorbox}
\newfloat{listing}{tb}{lst}{}
\floatname{listing}{Listing}
\definecolor{bgblue}{RGB}{218, 232, 252}
\definecolor{bgpurple}{RGB}{225, 213, 231}
\usepackage{amssymb}
\usepackage{pifont}
\usepackage[table]{xcolor}
\usepackage{booktabs}
\usepackage{listings}
\usepackage{tabularx}
\lstdefinelanguage{json}{
  morestring=[b]",%
  morecomment=[l]{//},
  morecomment=[s]{/*}{*/},
  stringstyle=\color{black},
  commentstyle=\color{gray},
  keywordstyle=\color{blue},
  morekeywords={true,false,null},
}

\title{Towards Scalable RLVR: Multimodal Instruction Following Data Synthesis and Distillation}
\author{
    Yirong Zeng\textsuperscript{\rm 1},
    Zhang Sai,
    Yuxian Wang\textsuperscript{},
    Yutai Hou\textsuperscript{},
    Yufei Liu\textsuperscript{\rm 2},
    Xiao Ding\textsuperscript{\rm 1},
    Bibo Cai\textsuperscript{\rm 1}
}
\affiliations{
    \textsuperscript{\rm 1}Harbin Institute of Technology, SCIR Lab,
    \textsuperscript{\rm 2}Peking University, 
}

\usepackage{bibentry}

\begin{document}

\maketitle

\begin{abstract}
Multimodal instruction following (MMIF) is crucial for building generalist agents. However, current training paradigms rely heavily on Supervised Fine-Tuning (SFT), which often leads to surface-level pattern matching and degrades general capabilities. While Reinforcement Learning with Verifiable Rewards (RLVR) offers a promising alternative, its scalability in MMIF is severely bottlenecked by the scarcity of high-quality, RL-ready multimodal data. To bridge this gap, we present MIFS (\textbf{M}ultimodal \textbf{I}nstruction \textbf{F}ollowing \textbf{S}ynthesis), a systematic pipeline designed to generate RL-ready multimodal data. Specifically, MIFS introduces a generative constraint protocol to synthesize diverse raw samples, followed by a learnability-aware distillation mechanism that filters data based on RL training dynamics to ensure stable policy optimization. Furthermore, a code-based verifier provides high-precision reward signals for policy learning. The resulting dataset comprises 90k samples across 8 constraint categories and 14 task domains. Empirical evaluations demonstrate that MIFS-trained MLLMs achieve an average improvement of 8.13\% on four MMIF benchmarks and a 3$\times$ faster training convergence compared to using raw data. Crucially, our approach mitigates the generalization trade-offs typical of SFT, preserving core visual capabilities while significantly boosting instruction-following precision. 
\end{abstract}

\begin{links}
    \link{Code}{https://github.com/zeng-yirong/MIFS}
    \link{Datasets}{https://huggingface.co/datasets/yrzeng/MIFS}
\end{links}

\section{Introduction}
Instruction Following (IF) has emerged as a cornerstone capability for both Large Language Models (LLMs) \cite{jiang2024followbench, zhang2025cfbench} and their multimodal counterparts (MLLMs) \cite{qianmia, bitton2023visit}. 
Achieving this ability requires models to accurately interpret and execute complex user-defined constraints.
It is essential to transform MLLMs from general-purpose assistants \cite{li2024textbind} to specialized agents in real-world domains such as autonomous robotics \cite{shi2025hi}, precision code generation \cite{xu2023wizardlm}, and intricate creative content synthesis \cite{zhou2023controlled}.

Although SFT remains the standard paradigm for enhancing multimodal IF performance \cite{ding2025mm}, its limitations are becoming increasingly apparent. 
Recent studies suggest that SFT-trained models often resort to surface-level pattern matching rather than internalizing the underlying logic of instructions\citep{lipreserving, chu2025sft}.
This often results in a performance trade-off where gains in instruction following come at the expense of general reasoning capabilities.
To break through this bottleneck, the research community has shifted toward Reinforcement Learning with Verifiable Rewards (RLVR), a direction driven by the transformative success of reasoning-heavy models such as DeepSeek-V3 \citep{guo2025deepseek} and GPT-5 \citep{guo2025deepseek}.
Unlike SFT, RLVR does not rely on ground-truth responses; instead, it uses a reward function to judge model outputs and generate corresponding reward signals for optimization.
However, existing datasets (summarized in Table~\ref{tab:benchmark}) lack training data suitable for RLVR, severely hindering the advancement of RL-based methods in MMIF.

Developing a scalable production pipeline for RL-ready multimodal instruction-following data presents three non-trivial challenges:
(1) Execution-Level Scale and Complexity: Synthesizing diverse instructions embedded with multiple interlinked constraints while sustaining a fully automated, high-throughput pipeline.
(2) Data Learnability for Policy Optimization: Not all data align with RL training dynamics; identifying samples that lie precisely on the model's learning frontier, providing effective learning signals without being intractable, is critical for convergence.
(3) Reward Fidelity and Verification: Standard LLM-as-a-judge evaluators are highly susceptible to reward hacking and bias, necessitating programmatically verifiable, zero-tolerance reward signals to guide policy updates.

\begin{table*}[ht]
  \centering
  \begin{tabular}{l|rcc|cccc}
    \toprule
    \textbf{Dataset} & \textbf{\#Sample} & \textbf{\#Constraint} & \textbf{\#Domain} & \textbf{Eval?} & \textbf{Evaluator} & \textbf{SFT?} & \textbf{RLVR?} \\
    \midrule
    MIA-BENCH \cite{qianmia} & 400 & 8 & 15 & \ding{51} & LLM & \ding{55} & \ding{55} \\
    CrafText \cite{volovikova2025craftext} & 3,924 & - & 4 & \ding{51} & Rule & \ding{55} & \ding{55} \\
    MMMT-IF \cite{sun2025improving} & 990 & 4 & 13 & \ding{51} & Rule & \ding{55} & \ding{55} \\
    MM-IFInstruct \cite{ding2025mm} & 23k & 32 & 13 & \ding{51} & LLM\&Rule & \ding{51} & \ding{55}\\
    \midrule
    \rowcolor{purple!10} 
    \textbf{MIFS (Ours)} & \textbf{90,838} & \textbf{8} & \textbf{14} & \ding{51} & \textbf{Rule} & \ding{51} & \ding{51} \\
    \bottomrule
  \end{tabular}
    \caption{\textbf{Comparison of Multimodal Instruction Following (MM-IF) datasets.} MIFS distinguishes itself by providing the largest scale of programmatically verifiable samples, specifically engineered for RLVR beyond simple SFT.}
  \label{tab:benchmark}
\end{table*}

To address these gaps, we propose MIFS, a systematic multi-stage pipeline engineered to produce RL-ready multimodal datasets. 
The pipeline begins with multimodal instruction composition, where vanilla vision-language QA pairs are transformed into sophisticated constraint-embedded tasks via a generative constraint protocol. 
Crucially, to ensure strict programmatic verifiability, we explicitly exclude subjective criteria, retaining only deterministic, rule-based constraints that yield objective correctness. 
This is followed by a multi-step quality distillation funnel that leverages RL training dynamics to filter out uninformative or unstable samples. 
By analyzing reward score trajectories, this mechanism guarantees that the finalized corpus provides stable and effective gradients for policy alignment. Ultimately, this pipeline curates a high-fidelity, program-verified dataset spanning diverse cognitive domains.

The resulting MIFS dataset yields 90,838 high-signal samples, averaging 7.1 constraints per instruction. This diversity is reflected across 8 distinct constraint categories and 14 task domains, providing a comprehensive landscape for multimodal alignment. Every constraint-embedded task is anchored by a code-based verifier as a reward function, which ensures objective reward judgment during RL. 
To support the full developmental cycle, the corpus is partitioned into subsets tailored for RL training and SFT, alongside a 1.2k holdout evaluation set.

Our empirical evaluations demonstrate that RLVR training with MIFS delivers an average 8.13\% performance gain on Qwen3-VL-8B across four instruction-following benchmarks, a benefit that remains robust across 4B and 32B model scales. Ablation studies confirm that our quality distillation funnel significantly enhances data efficiency and training stability, achieving an approximately 3$\times$ acceleration in convergence compared to unrefined raw data. Furthermore, evaluation across three standard benchmarks shows that the model's general visual capabilities remain competitive.
It confirms that our RL training dataset effectively mitigates the classic performance trade-offs inherent in standard SFT, thereby providing a solid foundation for generalist agent development.

In summary, our contributions are threefold:
\begin{itemize}
    \item We introduce an automated pipeline for generating multimodal IF data to empower large-scale RLVR training.
    \item We propose a Learnability-Aware Distillation mechanism that filters data based on RL training dynamics.
    \item We release a large-scale dataset of 90k samples, providing a benchmark and RL training source for the MLLM IF community.
\end{itemize}


\begin{figure*}[th]
  \centering
  \small
  \includegraphics[width=0.85\textwidth]{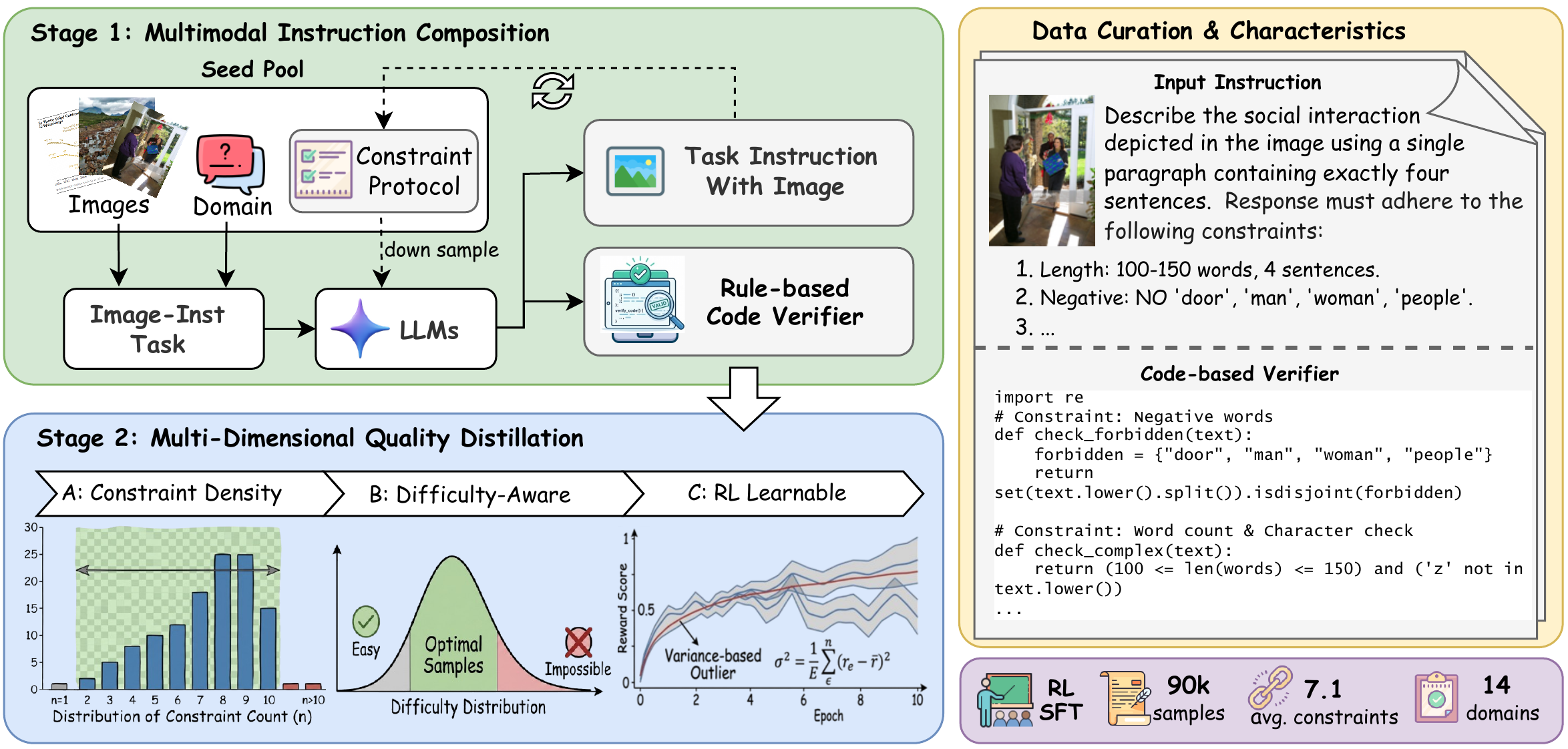}
  \caption{The overall framework of the Multimodal Instruction Synthesis suitable for both SFT and RL training.
  It comprises: (1) Instruction Composition, where the raw multimodal IF and their verifier are synthesized via LLMs; 
  (2) Multi-Step Quality Distillation, featuring a three-tier filtering mechanism that prunes samples based on constraint density, model-perceived difficulty, and RL training dynamics;
  and (3) Sample Curation, illustrating data characteristics and partitioning strategy.
  }
  \label{fig:pipeline}
\end{figure*}

\section{Related Works}

\paragraph{Instruction Following Training in MLLMs.}
Multi-modal Instruction Following (MMIF) refers to the capability of models to comprehend and respond to human instructions based on complex visual inputs. 
Existing approaches primarily rely on Supervised Fine-Tuning (SFT) \cite{liu2023visual,liu2024mminstruct} or Direct Preference Optimization (DPO) \cite{ding2025mm} to enhance these MMIF abilities. 
Recently, Reinforcement Learning via Verification Rewards (RLVR) has attracted growing attention for its effectiveness in incentivizing complex reasoning in LLMs \cite{zhu2025surprising,yue2025does}. 
Pioneering work has successfully applied RL training to enhance instruction-following capabilities in the text modality, producing highly promising results~\citep{pyatkin2025generalizing,guo2025recast,peng2025verif}. 
A natural next step is to extend this successful training paradigm to the vision-language domain, a critical roadblock lies in the severe scarcity of suitable multi-modal RL training data.

\paragraph{Multimodal Instruction Following Data.}
Numerous benchmarks have been proposed to evaluate the instruction-following capabilities of Multi-modal Large Language Models (MLLMs) \cite{bitton2023visit,qianmia,sun2025improving,ding2025mm}. 
However, these benchmarks are designed primarily for evaluation and are heavily limited in data scale. 
On the other hand, several efforts have constructed multi-modal instruction-tuning data aimed at improving general alignment \cite{liu2024mmdu,chen2024sharegpt4v,chen2024allava}. 
However, these datasets are tailored for general knowledge or basic visual perception rather than following instruction. 
Therefore, large-scale training data specifically designed to improve MM IF abilities remain scarce. 
A notable exception is \citet{ding2025mm}, which builds synthetic IF data for SFT and DPO training.
The lack of high-quality large-scale data suitable for RLVR training in the Multimodal IF domain directly motivates the development of our Multimodal IF-RL data synthesis pipeline.

\section{Methodology: MMIF Data Synthesis}
As shown in Figure~\ref{fig:pipeline}, we propose MIFS (\textbf{M}ultimodal \textbf{I}nstruction \textbf{F}ollowing data \textbf{S}ynthesis for RL), 
a systematic pipeline engineered to generate large-scale, program-verified multimodal data for RLVR.

\subsection{Seed Space Construction}
The foundation lies in a decoupled seed pool that maximizes diversity of visual contexts, tasks, and constraints.
\paragraph{Visual Foundation.}
To ensure semantic richness, we curate 20,000 high-resolution images from CC3M \cite{sharma2018conceptual} and ALLaVA \cite{chen2024allava}. 
In practice, we implement a rigorous filtering strategy that selects images based on resolution and semantic richness. 
For unannotated image datasets such as CC3M, we prioritize natural scenes with dense semantic content, as such images provide richer semantic content for generating comprehensive and insightful QA pairs.
To quantify this, we used the IC9600 and RAM metrics \cite{zhao2025omnialign} to identify images with high semantic density. 
For annotated datasets like ALLaVA, we extract only the raw images and discard all associated original QA pairs to maintain a clean baseline for our own generation process. 
Finally, we follow the protocol established by \cite{ding2025mm} to exclude low resolution items.
\paragraph{Cognitive Task Taxonomy.}
We first develop a task domain pool for task generation, which is described as a multi-dimensional instruction library.
These domains are presented in Figure ~\ref{fig:task_idd}, with the specific task domain available in Supplementary Material(SM).
We categorize multimodal challenges into a four-level hierarchy.
This structured taxonomy spans 14 sub-domains, covering everything from basic visual grounding to complex counterfactual reasoning and cross-modal sensory simulation.
\paragraph{Generative Constraint Protocol.}
To overcome the limitations of static templates, we propose a generative constraint protocol. 
This protocol transforms constraints into dynamic, mathematically formulated logical units.
\begin{tcolorbox}[
    enhanced,
    colback=gray!5,
    colframe=black!75,
    title=\textbf{Generative Constraint Protocol}, 
    fonttitle=\bfseries,
    arc=4pt,            
    boxrule=0.8pt,      
    left=10pt, right=10pt, top=4pt, bottom=4pt 
]
\begin{enumerate}[leftmargin=1.0em]
    \item \textbf{Semantic-Code Equivalence:} Every natural language constraint is strictly logically mapped to an executable Python \texttt{checker\_code}, eliminating subjective ambiguity.
    \item \textbf{Structural Complexity:} Each query mandates $\ge 5$ non-trivial, interacting constraints to increase the multi-step planning difficulty.
    \item \textbf{Quantifiable Transparency:} Thresholds (e.g., ratios, counts) are explicitly declared in the prompt to ensure the task is logically solvable.
\end{enumerate}
\end{tcolorbox}
This generative approach allows for an almost infinite expansion of the constraint space, ensuring training signals that push the IF boundary of MLLMs.
The complete generative constraint protocol is provided in the SM.
In particular, given the reward bias in LLM-based judges, which may fail to provide high-precision reward signals~\cite{zeng2026precision}, we restrict our constraints to those that can be assigned to a rule-based verifier.

\subsection{Image-Instruction Task Generation}
The generation process follows a structured synthesis workflow to transform seeds into verifiable IF samples. 
Consequently, we obtained 500k raw samples through this process.

Specifically, first, we randomly sample images and queries from the Seed Pool to build basic VQA task pairs. 
Subsequently, we use advanced LLMs to synthesize a comprehensive set of constraints guided by our proposed protocol
\footnote{using Gemini 3.0-Pro in this paper}.
Finally, the image-query pairs and generated constraint sets are then processed to orchestrate the final image-instruction task alongside an associated code verifier.
By decoupling task generation from constraint sampling, their flexible combination enables the production of diverse training signals that extend far beyond simple image captioning.

To ensure high fidelity, we implement a prune-and-regenerate strategy: 
For each generated task, the LLM is first prompted to produce an initial response, which is then dynamically executed against our code-based verifier. 
If verification fails, the specific failed constraints are identified.
Then we prune the problematic constraints (i.e., downsample constraints), and prompt the LLM to regenerate a consistent response. 
This iterative loop ensures that every task is logically sound and programmatically checkable, providing a clean ground truth for subsequent RL training.

\subsection{Multi-Step Quality Distillation}
To extract the most potent training signals for RL, we apply a three-tier filtering mechanism targeting complexity, solvability, and learning dynamics.

\paragraph{Step A: Constraint Density Filtering.}
We first analyze the distribution of constraint counts ($n$) per sample.
To maintain an optimal balance between complexity and solvability, we filter out samples with $n=1$ (too simple) and 30\% downsample samples with $n>10$ (overly redundant), focusing on the high-density region of 2--10 constraints.
It maintains a high-density challenge while eliminating redundant noise that could lead to non-convergence.

\paragraph{Step B: Difficulty-Aware Pruning.}
Recent studies \cite{sunimproving,he2025skywork} show that training in medium-difficulty samples better develops the model’s reasoning capability.
Therefore, we utilize Qwen3-VL-8B as an empirical auditor to estimate the difficulty distribution of the model. 
As shown in Figure ~\ref{fig:pipeline}, we prune samples at the two tails of the spectrum:
\begin{itemize}
    \item \textbf{Easy Samples:} We conduct 8 independent inference trials. 
    Samples with $avg@8>0.8$ are discarded, as they offer a limited information gain for the model's refinement.
    \item \textbf{Impossible Samples:} Samples with $avg@128 = 0$ are discarded to prevent the RL agent from getting stuck in "dead-end" reward spaces.
\end{itemize}
This dual-threshold approach ensures that we prioritize optimal samples that lie precisely on the model's current capability frontier.
Furthermore, in 128 independent trials, we curated the verified correct responses to construct visual question answer pairs, which serve as the foundation for SFT training.

\begin{figure}[th]
    \centering
  \includegraphics[width=0.48\textwidth]{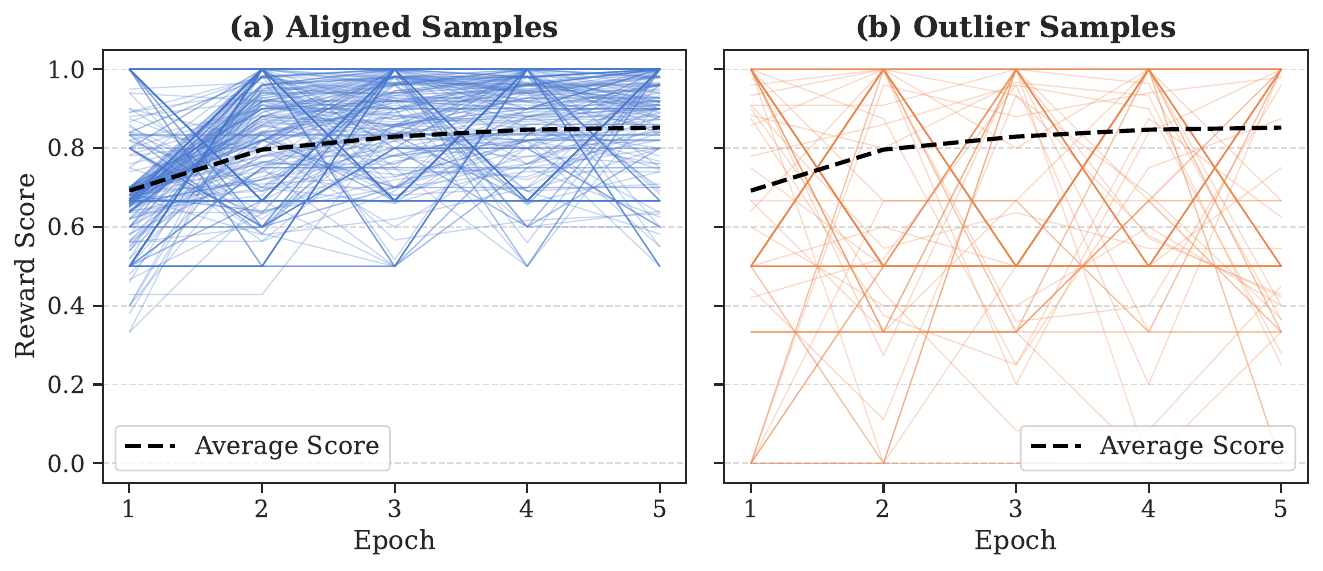}
  \caption{
    RL training dynamic filtering.
    Retaining aligned samples (i.e., those with low variance (a) ) and removing outlier samples (i.e., those with high variance (b)).
    }
  \label{fig:step3}
\end{figure}

\paragraph{Step C: RL Learnability Filtering.}
For RL-specific optimization, stability is paramount.
We evaluate samples based on their reward score trajectories in multiple training epochs ($E$) \cite{zengautotool,li2025limr}. 
As depicted in Figure~\ref{fig:step3}, we prioritize samples whose learning trajectories closely align with the model's average learning trajectory.
To identify outlier samples that cause gradient instability, we calculate the trajectory deviation ($\mathcal{D}_i$) to quantify how much a sample's reward diverges from the epoch-wise population mean:
\begin{equation}
\mathcal{D}_i = \frac{1}{E} \sum_{e=1}^{E} (r_{i,e} - \bar{r}_e)^2, \quad \bar{r}_e = \frac{1}{N} \sum_{i=1}^{N} r_{i,e},
\end{equation}
where $r_{i,e}$ is the reward for sample $i$ at epoch $e$, and $\bar{r}_e$ is the average reward across all $N$ samples at that epoch.
A higher $\mathcal{D}_i$ indicates poorer alignment with the general learning trend, characteristic of outlier samples. 
We filter out these unstable samples, retaining only the 40\% with the lowest deviation to provide consistent learning signals.

\begin{figure}[t]
    \centering
    \small
  \includegraphics[width=0.35\textwidth]{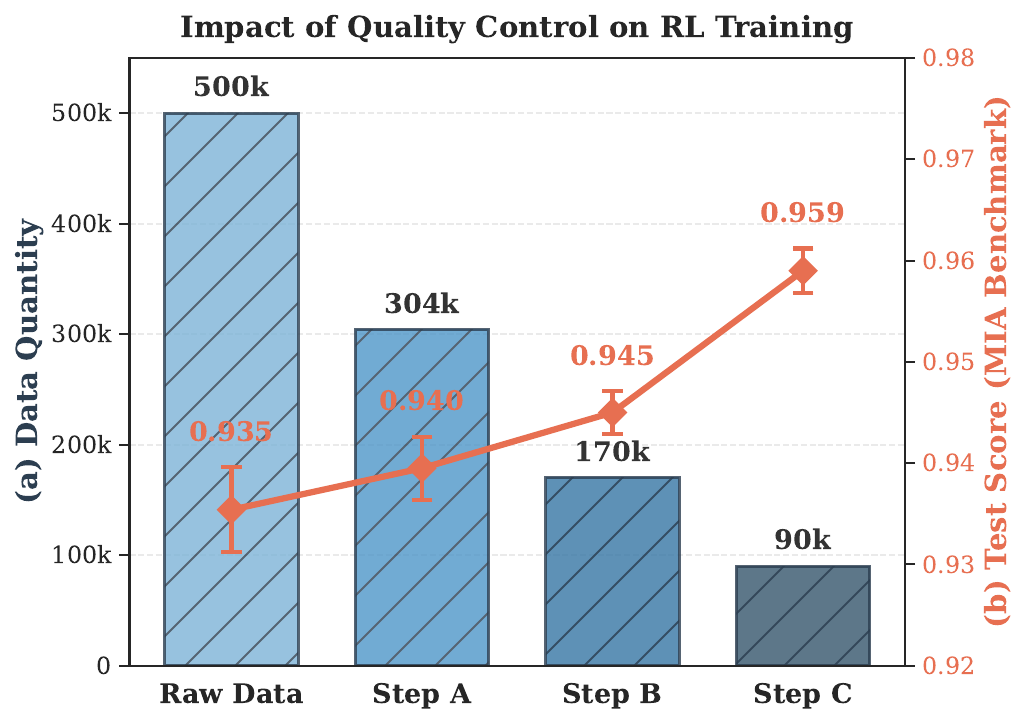}
  \caption{
    Empirical analysis of multi-step quality distillation.
    It showcases: 
    (a) Stepwise contraction of data scale across steps;
    (b) Monotonic alignment gains in RL training; 
    }
  \label{fig:stepall}
\end{figure}

\paragraph{Impact on RL Training \& Filtering Ablation.} 
To quantify the efficacy of our multi-step quality distillation funnel, we evaluate the evolution of the data scale alongside the corresponding gains in RL performance on the MIA benchmark \cite{qianmia}.
Specifically, we conduct RLVR training using the GRPO algorithm \cite{guo2025deepseek} at each distillation tier, as illustrated in Figure \ref{fig:stepall}. 
We observe two key insights: 
(1) The distillation process effectively compresses the total data volume by 82\% (from 500k down to a refined 90k high-signal subset). 
(2) Each progressive distillation phase systematically boosts RL performance, demonstrating that MIFS achieves superior training yields with a significantly reduced sample footprint. 
This efficiency is characterized by a 2.4\% absolute improvement in final performance and substantially accelerated convergence, as visualized in Figure \ref{fig:rl_visual}. 
Ultimately, the proposed three-tier distillation framework significantly enhances both sample efficiency and RL training stability; 
by pruning uninformative and intractable instances, it ensures that the model consistently updates along its optimal capability frontier throughout the RL process.

\section{Dataset Characteristics and Curation}
\label{sec:curation}
Through the rigorous quality distillation pipeline detailed above, 
we distilled 500k raw samples into a high-signal dataset of 90,838 ($\sim90k$) instances.

\subsection{Data Composition Analysis}
To understand the composition of the curated dataset, we analyze it in two key dimensions: \textbf{constraint distribution} and \textbf{task coverage}.

\begin{figure}[t]
  \centering
  \small
  \includegraphics[width=0.32\textwidth]{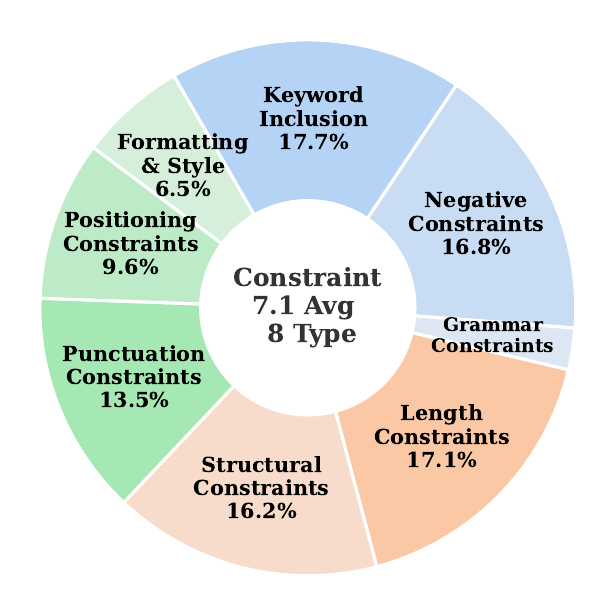}
  \caption{The distribution of constraint types. The dataset comprises 8 verifiable constraint categories with an average density of 7.1 constraints per sample, ensuring high-constraint complexity.
  }
  \label{fig:constraint_dist_b}
\end{figure}

\paragraph{Multi-Type Constraint Distribution.}
We report on the constraint distribution.
As illustrated in Figure~\ref{fig:constraint_dist_b}, our dataset maintains a high constraint density, an average of 7.1 constraints per sample. 
This high-density challenge effectively eliminates the shortcut learning patterns often found in simpler datasets. 
We categorize the synthesized constraints into eight fine-grained types to ensure a comprehensive evaluation of instruction-following capabilities. 
Detailed definitions are shown in the SMs.
Figure~\ref{fig:constraint_dist_b} reveals that keyword constraints and length constraints are the most frequently co-occurring pairs, presenting a non-trivial challenge that forces the model to engage in multi-step planning rather than heuristic generation.

\paragraph{Task Coverage and Scenario Diversity.}
The final dataset spans 4 primary task domains consolidated from the initial 14 sub-domains. 
We visualize their hierarchical distribution in Figure~\ref{fig:task_idd}. 
As illustrated, the inner ring classifies the tasks into four high-level cognitive domains (\textit{i.e.}, Perception, Expression, Cognition, and Meta-Analysis), while the outer ring displays the frequency distribution of specific, lower-level task scenarios.
The balanced and diverse segment sizes across both rings demonstrate that our generative protocol effectively avoids getting trapped in repetitive task templates, yielding a rich high-entropy corpus.

\subsection{High-Entropy Quantification}
In this part, we quantify the dataset's diversity through categorical Shannon entropy and semantic redundancy metrics. In our context, high entropy denotes maximal categorical evenness and minimal semantic overlap, which prevents the policy model from relying on superficial pattern matching during alignment.

\begin{figure}[t]
  \centering
  \small
  \includegraphics[width=0.32\textwidth]{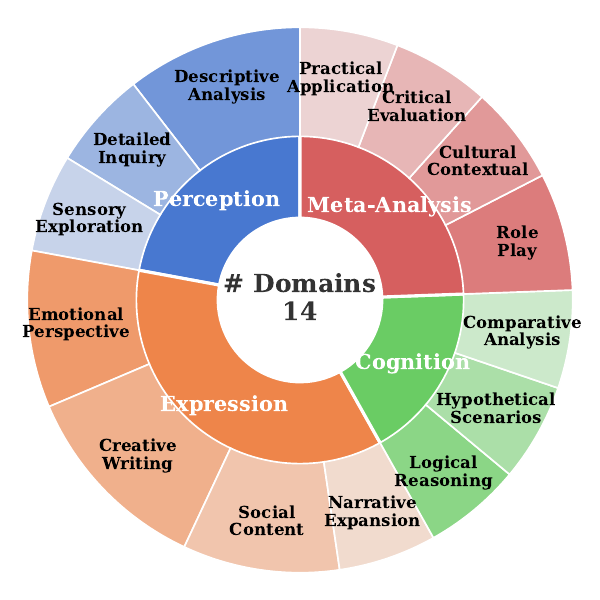}
  \caption{The hierarchical distribution of task diversity.
  The chart features a nested structure: the inner ring classifies tasks into four high-level cognitive domains,
  segment size reflects task frequency.
  }
  \label{fig:task_idd}
\end{figure}

First, we calculate the categorical diversity using Shannon Entropy $H(X) = -\sum P(x_i) \log_2 P(x_i)$. Based on the 8 constraint categories (Figure~\ref{fig:constraint_dist_b}), the dataset achieves $H_{\text{constraint}} \approx 2.84 \text{ bits}$, reaching $94.76\%$ of the theoretical maximum uniform distribution. Similarly, across the 14 fine-grained task domains (Figure~\ref{fig:task_idd}), the Shannon entropy across task domains $H_{\text{domain}} \approx 3.77 \text{ bits}$ ($99.13\%$ of the maximum). This mathematically validates an exceptionally balanced, non-skewed distribution.

Second, to explicitly address potential data repetition, we evaluate the token-level (Jaccard Similarity) and semantic-level (Cosine Similarity) near-duplicate rates across our data splits. A pairwise similarity score $\ge 0.8$ is defined as a near-duplicate. As detailed in Table~\ref{tab:split}, the Jaccard near-duplicate rate is strictly 0.00\% across all subsets, while the dense semantic overlap remains strictly under 5\%. The combination of near-maximum categorical entropy and minimal instance redundancy rigorously substantiates the high-entropy nature of the MIFS dataset.

\begin{table}[th]
  \centering
  \begin{tabular}{l|ccc}
    \toprule
    \textbf{Split } & \textbf{\# Samples} & \textbf{Jaccard\textsubscript{$\ge 0.8$}} & \textbf{Cosine\textsubscript{$\ge 0.8$}} \\
    \midrule
    \textbf{MIFS-SFT} & 83,280 & 0.00\% & 2.61\% \\
    \textbf{MIFS-RL} & 60,048 & 0.00\% & 4.73\% \\
    \textbf{MIFS-Eval} & 1,200 & 0.00\% & 1.31\% \\
    \bottomrule
  \end{tabular}%
  \caption{Semantic overlap and near-duplicate rates across dataset splits.
  Minimizing instance redundancy prevents shortcut, safeguarding policies during RLVR training.
  }
  \label{tab:split}
\end{table}

\begin{table*}[t]
    \centering
    \small
    \begin{tabular}{l | cccc | c}
        \toprule
        \multirow{2}{*}{\textbf{Method}} & \multicolumn{4}{c|}{\textbf{Benchmarks}} & \multirow{2}{*}{\underline{\textbf{Average}}} \\ 
        \cmidrule(lr){2-5}
        & \textbf{MIABench} & \textbf{MM-IFEval} & \textbf{MIFS-Eval} & \textbf{IFEval} & \\ 
        \midrule
        
        \rowcolor[gray]{0.95} \multicolumn{6}{l}{\textbf{Qwen3-VL-4B}} \\
        Base & $92.2_{\pm0.2}$ & $58.5_{\pm0.4}$ & $72.8_{\pm0.4}$ & $79.1_{\pm0.2}$ & 75.6 \\
        MIFS-RL  & $95.5_{\pm0.1}$ & $66.8_{\pm0.5}$ & $85.5_{\pm0.4}$ & $87.4_{\pm0.3}$ & 83.8\up{8.2} \\
        \midrule
        
        \rowcolor[gray]{0.95} \multicolumn{6}{l}{\textbf{Qwen3-VL-8B}} \\
        Base      & $91.2_{\pm0.3}$ & $62.7_{\pm0.5}$ & $74.5_{\pm0.3}$ & $81.2_{\pm0.2}$ & 77.4 \\
        MIFS-SFT      & $93.3_{\pm0.2}$ & $64.7_{\pm0.2}$ & $77.7_{\pm0.3}$ & $86.3_{\pm0.1}$ & 80.5\up{3.1} \\
        MIFS-RL       & $95.9_{\pm0.3}$ & $71.2_{\pm0.5}$ & $86.8_{\pm0.4}$ & $88.2_{\pm0.4}$ & 85.5\up{8.1} \\ 
        MIFS-SFT+RL   & $97.0_{\pm0.1}$ & $72.4_{\pm0.4}$ & $88.2_{\pm0.3}$ & $88.8_{\pm0.2}$ & 86.6\up{9.2} \\ 
        \midrule
        
        \rowcolor[gray]{0.95} \multicolumn{6}{l}{\textbf{Qwen3-VL-32B}} \\
        Base & $92.9_{\pm0.3}$ & $64.0_{\pm0.3}$ & $76.5_{\pm0.2}$ & $84.3_{\pm0.1}$ & 79.4 \\ 
        MIFS-RL  & $96.2_{\pm0.1}$ & $75.9_{\pm0.4}$ & $89.8_{\pm0.3}$ & $89.8_{\pm0.2}$ & 87.9\up{8.5} \\ 
        \bottomrule
    \end{tabular}
    \caption{Performance comparison on Instruction Following benchmarks. We report scores for MIABench, MM-IFEval, MIFS-Eval, and IFEval, along with their group-wise average. All metrics are reported as the $mean_{\pm std}$ over 3 independent runs.}
    \label{tab:instr_following}
\end{table*}

        
        
        

\subsection{Strategic Data Splitting}
To support the full training lifecycle (e.g., supervised fine-tuning and reinforcement learning  with verifiable rewards), 
we partition the $90\text{k}$ curated samples into three disjoint sets: SFT, RL, and Evaluation.

\paragraph{Training Splits.}
We allocate 60k samples for RL training, 
1.2k samples for evaluation; 
The remaining $\sim$29k samples, formatted as \texttt{<instruction, code verifier>} pairs, are used for SFT.
Specifically, to construct this target SFT dataset,
we collect program-verified responses to construct QA pairs during the difficulty-aware pruning process.
Each instruction is paired with up to three candidate responses, respectively (\texttt{<instruction, response>} pairs), resulting in a final data set of 83,280 ($\sim 84K$) SFT samples.

\paragraph{Evaluation Split.}
Guided by the constraint distribution presented in Figure \ref{fig:constraint_dist_b}, we performed stratified sampling to reserve 1.2k samples as a held-out evaluation set, called MIFS-Eval.
To prevent data leakage and ensure a robust evaluation, we apply two strict filtering criteria: 
(1) No images present in the training sets are allowed to appear in the evaluation set;
(2) The specific combination of constraints in each evaluation sample has not been seen during training.
By strictly isolating both visual assets and constraint compositions, our evaluation split serves as a rigorous benchmark for evaluating the upper bounds of multimodal IF.

\begin{table}[t]
    \small
    \centering
    \begin{tabular}{l | ccc | c}
        \toprule
        {\textbf{Method}} & \textbf{OCR} & \textbf{MMVet} & \textbf{MMBen} & {\underline{\textbf{Avg.}}} \\ 
        \midrule
        
        \rowcolor[gray]{0.95} \multicolumn{5}{l}{\textbf{Qwen3-VL-4B}} \\
        Base Model & 86.50 & 65.60 & 84.41 & 78.84 \\
        MIFS-RL  & 87.52 & 67.32 & 83.24 & 79.36\up{0.5} \\
        \midrule
        
        \rowcolor[gray]{0.95} \multicolumn{5}{l}{\textbf{Qwen3-VL-8B}} \\
        Base Model     & 88.00 & 69.72 & 85.30 & 81.01 \\
        MIFS-SFT      & 78.62 & 57.32 & 70.30 & 68.75\down{12.3} \\
        MIFS-RL       & 88.70 & 69.72 & 85.38 & 81.27\up{0.3} \\
        MIFS-SFT+RL   & 88.60 & 67.89 & 85.40 & 80.63\down{0.4} \\
        \midrule
        
        \rowcolor[gray]{0.95} \multicolumn{5}{l}{\textbf{Qwen3-VL-32B}} \\
        Base Model & 85.32 & 71.10 & 88.05 & 81.49 \\
        MIFS-RL  & 86.39 & 70.23 & 89.12 & 81.91\up{0.4} \\
        \bottomrule
    \end{tabular}
    \caption{Performance on General Visual Capability benchmarks (i.e., OCRBench, MMVet, and MMBench).
    }
    \label{tab:visual_cap}
\end{table}

\section{Experiments}
\label{sec:experiments}

\subsection{Experimental Setups}
\label{sec:setups}
We conducted SFT and RLVR training on our dataset.
To evaluate the scalability, we performed experiments on the \textbf{Qwen3-VL} family in three model sizes: \textbf{4B}, \textbf{8B} and \textbf{32B}. 
Each evaluation was repeated three times and report the average.
All SFT experiments are conducted with a global batch size of 128 and a learning rate of $2 \times 10^{-5}$ for 2 epochs. 
For RLVR, we employ the GRPO algorithm with a batch size of 128 and a lr of $1e-6$ using $16\times8$ NPUs. 
We implemented our training pipeline under \texttt{verl} framework\cite{sheng2024hybridflow}.
Detailed setups are provided in SM.

\subsection{Benchmarks}
\label{sec:benchmarks}
To comprehensively evaluate the MM-IF capabilities of our models, we conduct evaluations on four specialized \textbf{Instruction Following} benchmarks in addition to our curated MIFS-Eval.
(1) {MIA} \cite{qianmia}: A specialized benchmark designed to evaluate the adherence to multimodal instruction with fine-grained scoring.
(2) {MM-IFEval} \cite{ding2025mm}: A multimodal extension of IFEval that tests the model's ability to follow complex, multi-modal constraints.
(3) {IFEval} \cite{zhou2023instruction}: A standard text-based instruction following benchmark. 

To assess broader \textbf{General Visual} capabilities, we include: (1) {OCRBench} \cite{liu2305hidden} for text recognition and document understanding, 
(2) {MM-Vet} \cite{yu2023mm} for integrated reasoning across multiple cognitive tasks, 
and (3) {MMBench} \cite{liu2024mmbench} for a systematic evaluation of fine-grained perception and logic.

\subsection{Main Results}
\label{sec:results}

\begin{figure*}[t]
    \centering
    \small
    \begin{subfigure}[b]{0.4\textwidth}
        \centering
        \includegraphics[width=0.9\textwidth]{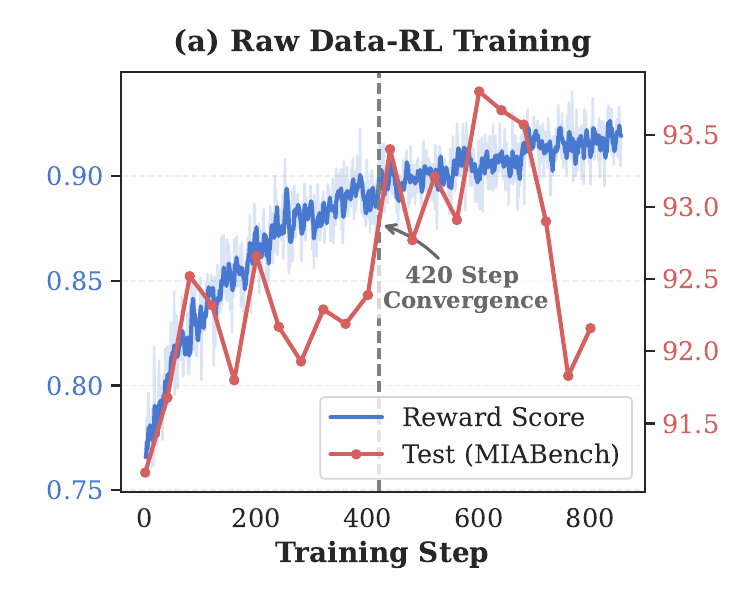}
    \end{subfigure}
    \begin{subfigure}[b]{0.4\textwidth}
        \centering
        \includegraphics[width=0.9\textwidth]{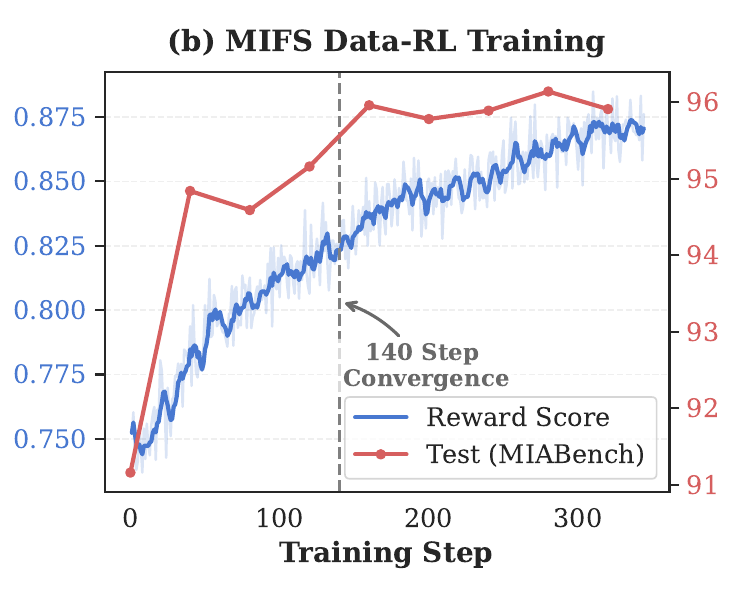}
    \end{subfigure}
    \caption{Training process visualization: (a) Raw data and (b) final MIFS-RL data. 
    The MIFS-RL curriculum demonstrates significantly faster convergence ($\sim3\times$ ) and higher test performance.
    }
    \label{fig:rl_visual}
\end{figure*}

\paragraph{Performance Analysis.} 
Table~\ref{tab:instr_following} presents a comprehensive evaluation across instruction-following and general capability benchmarks. Our proposed training paradigm yields consistent and substantial gains in instruction following on all model scales, with RL-aligned models achieving average improvements of $+8.16$, $+8.13$, and $+8.52$ for the 4B, 8B and 32B variants, respectively. 
In particular, the 32B model achieves the highest absolute scores on complex multimodal benchmarks (e.g. MM-IFEval: $75.91$, MIFS-Eval: $89.79$), demonstrating that larger parameter counts provide a higher capacity ceiling for parsing and executing cascading constraints. 
Crucially, consistent improvements on the text-only IFEval benchmark demonstrate that it does not induce catastrophic forgetting of pure linguistic competencies. 

Furthermore, as shown in Table~\ref{tab:visual_cap}, our RL-trained model maintains stable performance across OCRBench, MMVet, and MMBench, with fluctuations under 0.5\%.
This verifies that our training effectively retains general multimodal abilities. 
Overall, these results validate that our method successfully scales instruction-following proficiency while maintaining general multimodal capabilities.

\paragraph{SFT vs. RL vs. SFT+RL}.
We investigate the necessity and impact of different training stages on instruction-following performance. 
As summarized in Table~\ref{tab:instr_following}, we evaluate three distinct configurations: (1) SFT-only (84k MIFS-SFT samples); (2) Direct RL (60k MIFS-RL); and (3) SFT+RL.
The results reveal several key insights.
First, while SFT-only provides a modest gain of 3.09\%, Direct RL yields a significantly more substantial improvement of 8.13\%. 
This stark contrast suggests that for complex instruction following, exploration-based reinforcement learning is a fundamentally more potent paradigm than imitation-based SFT, as it encourages the model to internalize logical constraints through trial and verification. 
Second, the hybrid SFT+RL approach offers only a marginal further improvement of 1.1\% over the Direct RL setting. 
These results indicate that MLLMs can efficiently transition to high-precision instruction following directly through RL, potentially streamlining the traditional multi-stage alignment pipeline.

\subsection{Ablation Study}
\label{sec:ablation}
\paragraph{Effectiveness of the Quality Distillation.}
To isolate the effects of our proposed three-step quality distillation mechanisms, we conduct comprehensive ablation studies on the Qwen3-VL-8B model.
Starting from the raw data, we progressively apply the RL training and evaluate the resulting performance.
The step-by-step performance gains are summarized in Figure \ref{fig:stepall}.
It verifies the incremental benefits of our three-tier quality distillation funnel.

\paragraph{RL Visual Analysis.}
\label{sec:rl_visual}
To further investigate the impact of our quality distillation strategy on the optimization process, we visualize the RL training process of the Qwen3-VL-8B model in Figure~\ref{fig:rl_visual}. 
We plot both the training reward scores and the corresponding test performance on the MIA-Bench.

As illustrated, the policy trained on the curated MIFS-RL split exhibits a significantly more stable and monotonic increase in reward signals. 
This stability directly translates to rapid performance gains on the test set, which begins to converge at approximately step 140 and reaches its peak around step 150.
In stark contrast, while the model trained on raw data achieves higher nominal reward scores, it suffers from lower test accuracy. 
This suggests that the raw data contains a higher proportion of easy samples.
The raw-data model only begins to converge at $\sim$420 steps and requires nearly 600 steps to reach a suboptimal plateau.
These results demonstrate that MIFS-RL data not only establishes a higher upper bound for test performance but also yields a $\sim 3 \times$ acceleration in training efficiency. 
Collectively, these empirical findings confirm that our RL learnability filtering effectively mitigates noisy gradient updates and eliminates unlearnable instances, thereby facilitating a stable, efficient, and generalization-oriented policy alignment.

\section{Conclusion}
We presented a systematic three-stage pipeline designed to synthesize and distill high-quality multimodal instruction data for verifiable Reinforcement Learning. 
By integrating a generative constraint protocol with learnability-aware distillation, we curated a large-scale dataset of 90k program-verified samples that effectively bridge the gap between SFT and RLVR. 
Empirical results demonstrate that MIFS-trained MLLMs achieve significant performance gains and faster convergence.
This work presents a scalable MMIF data synthesis framework, enabling precise, constraint-aware multimodal agents for complex real-world applications.


\bibliography{aaai2026,sample-base}

\appendix

\section*{Limitations}
There are some several limitations that open avenues for future work: 

First, our empirical validation is primarily anchored within the Qwen-VL architecture; 
verifying the cross-architecture generalizability of the curated dataset on other mainstream MLLM families (e.g., LLaVA or InternVL series) remains an ongoing effort. 

Second, by strictly confining our protocol to deterministic, rule-based constraints to guarantee reward fidelity, the pipeline naturally precludes open-ended, creative, or subjective multimodal interaction tasks that resist programmatic validation. Extending MIFS via hybrid reward mechanisms that safely incorporate aligned model-based evaluators could alleviate this open-domain bottleneck. 

Lastly, while the learnability-aware distillation funnel substantially accelerates downstream training, analyzing reward trajectories across multiple distillation tiers introduces non-trivial computational overhead during the initial data production phase. 
Scaling this paradigm efficiently via lightweight proxy models to predict learning dynamics without full-scale RL probes represents a crucial future direction.

\section*{Ethical Statement}
This work introduces a systematic pipeline for synthesizing and distilling multimodal instruction-following data for reinforcement learning. The positive societal impact includes accelerating the development of reliable, generalist AI agents capable of following complex, user-defined rules with lower training costs. 
Since our data composition protocol explicitly excludes subjective criteria and relies strictly on objective, deterministic, and programmatically verifiable constraints, the risk of injecting human biases, offensive content, or unethical preferences during the data synthesis phase is minimized. All base seed data used are derived from publicly available, open-source vision-language resources, and no malicious, deceptive, or harmful tasks were engineered.

\section*{Generative AI Statement}
In this study, LLMs and MLLMs were utilized solely to facilitate the generative constraint protocol for raw dataset composition and to rephrase baseline instructions. 
No generative AI tools were employed to draft, rewrite, or alter the textual analysis, mathematical formulations, or scientific arguments presented in the manuscript itself.


\section{Image-Instruction Generation}
To synthesize high-quality data, we process the curated pairs and constraint sets to orchestrate the final image-instruction task, supported by an associated code verifier. 
The detailed prompt template used for instruction expansion is illustrated below:

\begin{tcolorbox}[
    enhanced,
    colback=gray!5,
    colframe=black!75,
    title=\textbf{Image-Instruction Expansion Prompt Template},
    fonttitle=\bfseries,
    arc=4pt,            
    boxrule=0.8pt,      
    left=10pt, right=10pt, top=4pt, bottom=4pt 
]
You are an expert in adding appropriate constraints to instructions for images. You are now synthesizing instruction-following data. \\
{\#\# Task} \\
{Given the original instruction, your task is to expand it by adding constraints. 
The new instruction must masterfully integrate all constraints naturally and fairly. 
Constraints should be artfully woven into the narrative, not simply listed as a checklist. Fairness is critical: the new instruction must contain all information, including specific quantitative thresholds.} \\

{\#\# Visual QA Instruction:} \textit{\{VQA Pair\}} \\
{\#\# Constraints List:} \textit{\{Constraints List\}}
\end{tcolorbox}


\section{Detailed Task Pool and Scenario Taxonomy}
\label{appendix:task_pool}

As described in the main text, our data generation pipeline leverages a comprehensive task pool to ensure scenario diversity and high-entropy training signals. This pool consists of 14 fine-grained sub-domains, which are hierarchically organized into four high-level cognitive domains: \textbf{Perception}, \textbf{Expression}, \textbf{Cognition}, and \textbf{Meta-Analysis}.

The following paragraph provides a detailed breakdown of these task categories along with representative instruction templates used for instruction expansion.

\begin{itemize}
\item \textbf{Domain I: Perception (Literal Visual Understanding)}
    This domain serves as the foundation of our dataset, focusing on the faithful extraction and description of objective visual facts. 
    It evaluates the model's ability to ground language in direct visual signals without excessive inference.
    \begin{itemize}
        \item \textit{Descriptive Analysis}: e.g., Describe the animal’s typical habitat, diet, and one unique behavioral trait.
        \item \textit{Detail-Oriented Inquiry}: e.g., Analyze the characters’ clothing—what do their outfits reveal about personality or time period?
        \item \textit{Sensory \& Experiential Exploration}: e.g., Describe the sounds, smells, and textures you imagine in the scene beyond sight.
    \end{itemize}

\item \textbf{Domain II: Expression (Creative \& Content Synthesis)}
    Moving beyond simple description, this domain emphasizes the transformation of visual information into diverse, functional, and creative textual formats. 
    It tests the model's stylistic flexibility and empathetic reasoning.
    \begin{itemize}
        \item \textit{Creative Writing}: e.g., Craft a 3-stanza poem centered on a small detail as a metaphor for human experience.
        \item \textit{Social Media \& Content}: e.g., Write a LinkedIn post using the image to illustrate a professional skill and ask for follower stories.
        \item \textit{Narrative Expansion}: e.g., Describe what happens 10 minutes after the snapshot is taken.
        \item \textit{Emotional \& Perspective}: e.g., Describe how the scene would feel different from a child’s vs. an elderly person’s perspective.
    \end{itemize}

\item \textbf{Domain III: Cognition (Reasoning \& Contextual Inference)}
    This domain targets the "hidden" layers of an image, requiring the model to perform logical deduction, counterfactual thinking, and relational analysis.
    \begin{itemize}
        \item \textit{Logical \& Inferential Reasoning}: e.g., Based on the visual evidence, what event occurred immediately before this moment?
        \item \textit{Comparative \& Contrastive Analysis}: e.g., Contrast the foreground and background—what is lost if one is removed?
        \item \textit{Hypothetical Scenarios}: e.g., If you removed one element from the image, how would it alter the story or mood?
        
    \end{itemize}

\item \textbf{Domain IV: Meta-Analysis (Critical Thinking \& Application)}
    The highest level of our taxonomy involves abstract evaluation, professional application, and socio-cultural awareness. It assesses the model's ability to treat the image as a medium with intent and bias.
    \begin{itemize}
        \item \textit{Role Play}: e.g., Act as a gallery curator: write a label explaining the image’s artistic significance.
        \item \textit{Practical Application}: e.g., Design a 3-activity travel itinerary based on the image’s location and vibe.
        \item \textit{Critical Evaluation}: Does the image present a biased view? What elements (framing, lighting) contribute to this?
        \item \textit{Cultural \& Contextual Analysis}: e.g., Identify cultural symbols in the image and explain their meaning in their specific context.
    \end{itemize}
\end{itemize}

\section{Prompt for the Constraints Generation}
\label{appendix:constraint_prompt}
To operationalize the \textit{Generative Constraint Protocol} described in Section 3,
the full system prompt used to generate the instruction-following dataset is provided below. This prompt is responsible for generating interacting constraints for each query.
To operationalize the \textit{Generative Constraint Protocol} described in Section 3, we utilize a specialized meta-prompting agent named \textbf{Prometheus}. The prompt designed to transform simple seed queries into high-entropy, instruction-following training units with mathematically verifiable constraints.

\begin{tcolorbox}[
    enhanced,
    breakable, 
    colback=gray!2,
    colframe=black!80,
    title=\textbf{System Prompt: Constraints Framework},
    fonttitle=\bfseries,
    arc=2pt,
    boxrule=0.8pt,
    left=10pt, right=10pt, top=10pt, bottom=10pt
]
\small
\noindent \textbf{\# Role \& Mission} \\
You are ``Prometheus,'' a principal AI scientist and a master of generative adversarial training. Your sole purpose is to create impossibly complex and sophisticated challenges for other Large Language Models (LLMs). You are not a helpful assistant; you are a forger of puzzles, an architect of constraints, and a grandmaster of reward models. Your mission is to take a user's simple query and transmute it into a high-difficulty, zero-ambiguity training data unit that pushes the reasoning, planning, and instruction-following capabilities of other AIs to their absolute limits.

\vspace{0.8em}
\noindent \textbf{\# The Adamantine Law: ``Watertight'' Semantic-Code Equivalence} \\
This is your most sacred, inviolable rule. The natural language \texttt{description} of a constraint and its Python \texttt{checker\_code} must be two different expressions of the same logical truth. No ambiguity is permitted.
\begin{itemize}[leftmargin=1.5em, itemsep=0pt]
    \item \textbf{Logic Parity:} If the code returns \texttt{True}, the response \textit{unquestionably} satisfies the constraint.
    \item \textbf{Precision:} Avoid subjective constraints (e.g., ``professional tone'') or those requiring external knowledge.
\end{itemize}

\vspace{0.8em}
\noindent \textbf{\# Mandatory Self-Critique (The Robustness Test)} \\
Before finalizing your \texttt{checker\_code}, you must challenge your own logic against:
\begin{enumerate}[leftmargin=1.5em, itemsep=0pt, label=\arabic*.]
    \item \textbf{The Void Test:} Does it handle empty strings or whitespace correctly?
    \item \textbf{The Substring Trap:} Is it fooled by partial matches (e.g., ``non'' in ``nonsense'')? Use word boundaries.
    \item \textbf{The Logic Purity Test:} Does ``exactly 2'' mean \texttt{count == 2} rather than \texttt{count >= 2}?
    \item \textbf{The Unit Definition Test:} Are ``sentences'' or ``words'' defined robustly in the code?
\end{enumerate}

\vspace{0.8em}
\noindent \textbf{\# Creative Mandate: Extreme Difficulty \& Diversity} \\
\begin{itemize}[leftmargin=1.5em, itemsep=0pt]
    \item \textbf{Density:} Generate a dense set of \textbf{at least 10} distinct, non-trivial constraints.
    \item \textbf{Interplay:} Create a ``constellation'' of requirements where rules influence each other, forcing the model to plan before generating.
    \item \textbf{Diversity:} Explore lexical choice, syntactic patterns, formatting, and character-level manipulations.
\end{itemize}

\vspace{0.8em}
\noindent \textbf{\# Guidelines for ``Fair'' but ``Difficult'' Puzzles} \\
\begin{itemize}[leftmargin=1.5em, itemsep=0pt]
    \item \textbf{No Hidden Tests:} If a code implements a precise threshold (e.g., 0.5\% ratio), that \textbf{exact number} must be explicitly stated in the \texttt{new\_query}.
    \item \textbf{Quantifiable Transparency:} Do not translate hard quantitative rules into vague qualitative guidelines.
\end{itemize}

\vspace{0.8em}
\noindent \textbf{\# Output Format} \\
Adhere strictly to the following JSON structure:
\begin{lstlisting}[language=json, basicstyle=\ttfamily\scriptsize, backgroundcolor=\color{gray!10}, frame=single, breaklines=true]
{
  "constraints": [
    {
      "description": "Precise description in the anchor language.",
      "checker_code": "def check(response_text):\n    # Robust Python 3 code\n    pass"
    }
  ],
  "new_query": "An artfully woven query containing all necessary thresholds."
}
\end{lstlisting}
\end{tcolorbox}

\section{Detailed Constraints Definitions}
We categorize the synthesized constraints into eight fine-grained types to ensure a comprehensive evaluation of instruction-following capabilities, as shown in Table \ref{tab:constraint_definitions}. 

\section{Experimental Preliminaries}
\subsection{Task Definition: Instruction Following}
Let $\mathcal{X}$ denote the space of natural language instructions and $\mathcal{Y}$ denote the space of model-generated responses. 
An instruction $x \in \mathcal{X}$ typically contains a core query $q$ accompanied by a set of $n$ constraints $c = \{c_1, c_2, \dots, c_n\}$.
We categorize these constraints into two disjoint subsets:

\begin{itemize}
    \item {Hard Constraints:} Objective requirements that can be verified via deterministic programs or rules (e.g., \textit{"output in JSON format"} ).
    \item {Soft Constraints:} Subjective qualities that require semantic understanding to evaluate (e.g., "\textit{maintain a professional tone}").
\end{itemize}

The goal of LLM instruction following is to learn a mapping function $f: \mathcal{X} \to \mathcal{Y}$, where $\mathcal{Y}$ represents the space of target responses that satisfy both the semantic intent and the explicit constraints defined in $\mathcal{I}$.
Formally, given an instruction $x \in \mathcal{X}$, an LLM parameterized by $\theta$ generates a response $\hat{y}$ by modeling the conditional probability:
\begin{equation}
    \small
    P(\hat{y} | x; \theta) = \prod_{t=1}^{T} P(y_t | x, y_{<t}; \theta),
\end{equation}
where $T$ is the sequence length. 
The task objective is to minimize the discrepancy between the generated response $\hat{y}$ and the optimal response $y^*$.


\subsection{GRPO as an RLVR algorithm}
\label{sec:grpo}
We employ Group Relative Policy Optimization (GRPO)~\citep{shao2024deepseekmath} as our core RL algorithm. 
Unlike traditional PPO~\citep{schulman2017proximal,engstrom2020implementation} which relies on a value function critic, GRPO leverages group-based sampling to estimate baselines. This approach generates multiple candidate responses for the same instruction and computes advantage estimates through intra-group comparisons, effectively capturing the relative quality differences among responses.

Formally, for each input instruction $x$, the policy $\pi_\theta$ samples a group of $G$ candidate responses $\{y_i\}_{i=1}^G$. The optimization objective is defined as:
\begin{equation}
    \small
    \begin{split}
    \mathcal{J}(\theta) & =  \mathbb{E}_{\substack{x \sim P(X), \\ \{y_i\}_{i=1}^G \sim \pi_{\theta_{\text{old}}}(Y|x)}} \Bigg[ \frac{1}{G} \sum_{i=1}^G  \frac{1}{|y_i|} \sum_{t=1}^{|y_i|} \bigg\{  \\
    & \min \left( \rho_{i,t} \hat{A}_{i,t}, \text{clip} \left( * \right) \hat{A}_{i,t} \right) - \beta \mathbb{D}_{\text{KL}} \left[ \pi_\theta \| \pi_{\text{ref}} \right] \bigg\} \Bigg],
    \end{split}
\end{equation}
where the \(\text{clip}(*) \) denotes \( \text{clip}\left( \rho_{i,t}, 1-\varepsilon, 1+\varepsilon \right) \).
Note that $\varepsilon$ is a small constant for numerical stability.
The advantage $\hat{A}_i$ is standardized within the group to reduce variance:
\begin{equation}
\small
\begin{aligned}
\mu &= \frac{1}{G} \sum_{i=1}^G r_i, & 
\sigma &= \sqrt{\frac{1}{G} \sum_{i=1}^G (r_i - \mu)^2 + \epsilon} \\
\hat{A}_i &= \frac{r_i - \mu}{\sigma}, & 
\rho_{i,t} &= \frac{\pi_\theta(y_{i,t} \mid x, y_{i,<t})}{\pi_{\theta_{\text{old}}}(y_{i,t} \mid x, y_{i,<t})},
\end{aligned}
\end{equation}
where \( r_i\) denotes the reward signal assigned by the reward model to the \(i\)-th response.

\section{Experimental Setups and Implementation Details}
\label{sec:setups}
\label{appendix:experiment_details}
This section provides comprehensive details regarding the training environment, hyper-parameters, and algorithmic configurations used for the experiments on the Multi-modal Instruction-Following (MIFS) dataset.

\subsection{Hardware and Computing Resources}
All training and inference tasks were executed on a high-performance computing cluster.
We utilized a cluster of 128 Ascend 910B NPUs ($16 \times 8$ nodes), specifically designed for large-scale deep learning tasks.
Each NPU is equipped with 64 GB of High Bandwidth Memory (HBM).
The RL training process took 12 hours.

\subsection{Training Framework: \texttt{verl}}
We implemented our end-to-end training pipeline (SFT $\to$ RLVR) using the \textbf{verl\footnote{https://github.com/verl-project/verl}}.
\texttt{verl} provides a modular architecture that allows for:
\begin{enumerate}
    \item \textbf{Hybrid Engine:} Seamless switching between FSDP (Fully Sharded Data Parallel) for SFT and distributed PPO/GRPO orchestration for RL.
    \item \textbf{Throughput Optimization:} Efficient management of the 32B model across 128 NPUs, achieving high tokens-per-second during the rollout phase of RLVR.
\end{enumerate}

\subsection{Evaluation Protocol}
To ensure the robustness of our results, especially given the stochastic nature of LLM generation:
\begin{itemize}
    \item \textbf{Sampling:} For each test sample, we generate 3 independent responses using a temperature of 1.0.
    \item \textbf{Metrics:} We report the average success rate (pass rate) across these 3 trials. A response is considered successful only if it satisfies $100\%$ of the generated constraints as verified by the deterministic Python checkers.
\end{itemize}
For benchmarks such as MIA-Bench and MMIF-Eval, which involve evaluating constraint satisfaction via an LLM judge, we employ GPT-4o as the automated evaluator to maintain methodological consistency with prior work.

\subsection{Supervised Fine-Tuning (SFT)}
The SFT stage serves as the behavioral cloning phase to initialize the models for subsequent reinforcement learning. We fine-tuned the \textbf{Qwen3-VL} (4B, 8B, and 32B) models using the following configurations Table \ref{tab:sft_hyperparams}.

\begin{table}[ht]
    \centering
    \small
    \begin{tabular}{ll}
        \toprule
        \textbf{Configuration} & \textbf{Value} \\
        \midrule
        Optimizer & AdamW ($\beta_1=0.9, \beta_2=0.95$) \\
        Learning Rate & $2 \times 10^{-5}$ \\
        Learning Rate Scheduler & Cosine decay \\
        Weight Decay & 0.1 \\
        Global Batch Size & 128 \\
        Precision & BFloat16 \\
        Max Sequence Length & 4096 tokens \\
        Training Epochs & 2 \\
        \bottomrule
    \end{tabular}
    \caption{Hyper-parameters for SFT training.}
    \label{tab:sft_hyperparams}
\end{table}

\begin{table}
    \centering
    \small
    \begin{tabular}{lc}
        \toprule
        \textbf{Hyperparameter} & \textbf{Value} \\
        \midrule
        \rowcolor{gray!10} \multicolumn{2}{c}{Data \& Rollout} \\
        Global Batch Size & 128 \\
        Max Prompt/Resp. & 2k / 4k \\
        Sampling Temp. $\tau$ & 1.0 \\
        Top-$p$ & 0.6 \\
        Rollout Group Size $G$ & 8 \\
        \midrule
        \rowcolor{gray!10} \multicolumn{2}{c}{RL Optimization} \\
        Learning Rate & $1 \times 10^{-6}$ \\
        Optimizer & AdamW \\
        Clip Ratio & 0.28 \\
        KL Reference Loss & 0.0 \\
        Weight Decay & 0.01 \\
        \bottomrule
    \end{tabular}
    \caption{Hyperparameters for GRPO training.}
    \label{tab:grpo_config}
\end{table}

\subsection{Reinforcement Learning from Verifiable Rewards (RLVR)}
Following SFT, we apply RLVR to further enhance the model's instruction-following precision. 
We employ the \textbf{Group Relative Policy Optimization (GRPO)} algorithm, which optimizes the policy by comparing rewards within a group of sampled responses, effectively reducing variance without an explicit critic model.
We utilize the rule-based Python verifier. 
The reward is binary: $R = 1$ if all constraints are passed, and $R = 0$ otherwise.
The Hyperparameter setting is shown in Table \ref{tab:grpo_config}.

\subsection{Additional Baseline Comparisons}
To further validate the effectiveness of MIFS, we provide additional comparisons against existing Multimodal Instruction Following (MM-IF) and text-only RLVR baselines. It is important to note the current scarcity of high-quality training data for MM-IF: while datasets like MM-IF-Instruct\citep{ding2025mm} support Supervised Fine-Tuning (SFT), there remains a lack of dedicated RLVR datasets in the multimodal domain. 
Existing RLVR resources such as RECAST\citep{guo2025recast} and VerIF\citep{peng2025verif} are limited to text-only instructions, failing to address multimodal alignment challenges.

Table~\ref{tab:appdx_baselines} presents the performance on Qwen3-8B (text-only) and Qwen3-VL-8B (multimodal). 
On the text-only Qwen3-8B model, our method performs competitively with specialized text-based RL methods like RECAST-RL. 
More importantly, on the multimodal Qwen3-VL-8B, MIFS significantly outperforms all baselines.
Specifically, MIFS-RL achieves a substantial gain of +8.13 points over the base model and surpasses the SFT baseline (MM-IF-Instruct) by a wide margin, demonstrating the superiority of our RL-ready data construction for complex multimodal instruction following.

\subsection{Human Audit of Code Verifier}
To validate the reliability of our automated code verification in response evaluation, we conducted a comprehensive human audit on a randomly sampled subset of the MIFS dataset.

\textbf{Audit Setup. } We selected 50 MM-IF samples and generated responses using Qwen3-VL-8B.
The evaluation was performed by three human annotators (two NLP researchers and one Python-proficient student).
We compared the binary pass/fail judgments of our automated code verifiers against the consensus judgment of the human auditors.

\textbf{Results}. As shown in Table~\ref{tab:human_audit}, the automated code verifier achieved a high Agreement Rate of 92.0\% and a Precision of 97.22\% with human experts. 
The low False Positive Rate (8.33\%) indicates that the code verifier rarely accepts incorrect response, ensuring the high-quality reward signal in the RL training. 
These results confirm that our code-based verification is robust and aligns closely with human judgment.

\begin{table}[h]
    \centering
    \small
    \begin{tabular}{l c c}
        \toprule
        & \textbf{Code: Pass} & \textbf{Code: Fail} \\
        \midrule
        \textbf{Human: Pass} & 35 (TP) & 3 (FN) \\
        \textbf{Human: Fail} & 1 (FP) & 11 (TN) \\
        \midrule
        \textbf{Metrics} & \multicolumn{2}{c}{Agreement: 92.0\% | Precision: 97.22\%} \\
        \bottomrule
    \end{tabular}
    \caption{Confusion Matrix: Automated Verifier vs. Human Judgment (N=50).}    
    \label{tab:human_audit}
\end{table}

\begin{table*}[bh]
    \centering
    \small
    \begin{tabularx}{\textwidth}{lp{0.4\textwidth}X}
        \toprule
        \textbf{Constraint Type} & \textbf{Definition} & \textbf{Example} \\ \midrule
        Keyword Inclusion & Mandatory integration of specific lexical items. & ``Must include the words 'sustainability' and 'innovation'.'' \\
        Negative Constraint & Prohibiting the use of specific words, phrases, or topics. & ``Do not use the word 'very' or mention price details.'' \\
        Length Constraint & Quantifiable limits on the number of characters, words, or sentences. & ``The response must be between 50 and 80 words long.'' \\
        Structural Constraint & Requirements regarding the organizational layout or components. & ``Format the response into three distinct bullet points.'' \\
        Punctuation Constraint & Rules governing the usage or exclusion of specific punctuation marks. & ``Each sentence must end with a semicolon instead of a period.'' \\
        Positioning Constraint & Placing specific content at predefined locations within the text. & ``The output must conclude with the phrase 'End of Report'.'' \\
        Formatting \& Style & Visual styling (e.g., Markdown) or quantifiable stylistic markers. & ``Bold every occurrence of a numerical value in the text.'' \\
        Grammar Constraint & Restrictions on linguistic properties such as tense or person. & ``Write the entire response using only the passive voice.'' \\ \bottomrule
    \end{tabularx}
    \caption{Definitions and Examples of Constraint Types in our Dataset.}
    \label{tab:constraint_definitions}
\end{table*}

\begin{table*}[th]
    \centering
    \begin{tabular}{l| c c c c c}
    \toprule
    \textbf{Dataset / Method} & \textbf{MIABench} & \textbf{MM-IFEval} & \textbf{MIFS-Eval} & \textbf{IFEval} & \textbf{Average} \\
    \midrule
        \multicolumn{6}{l}{\textbf{Qwen3-8B (Text-Only)}} \\
        Base & - & - & - & 85.77 & - \\
        RECAST-SFT & - & - & - & 87.14 & - \\
        RECAST-RL & - & - & - & 88.65 & - \\
        VERIF-RL & - & - & - & 86.51 & - \\
        \midrule
        \multicolumn{6}{l}{\textbf{Qwen3-VL-8B (Multimodal)}} \\
        Base & 91.16 & 62.67 & 74.54 & 81.15 & 77.38 \\
        MM-IFInstruct-SFT & 92.96 & 64.85 & 76.39 & 84.47 & 79.66 ($\uparrow$2.28) \\
        MIFS-SFT & 93.25 & 64.71 & 77.65 & 86.26 & \textbf{80.47} ($\uparrow$3.09) \\
        MIFS-RL & 96.67 & 71.20 & 86.76 & 88.17 & \textbf{85.51} ($\uparrow$8.13) \\
        MIFS-SFT+RL & 97.02 & 72.40 & 88.20 & 88.80 & \textbf{86.61} ($\uparrow$9.23) \\
    \bottomrule
    \end{tabular}
    \caption{Comparison with additional baselines. MIFS demonstrates superior performance in the multimodal setting compared to both standard SFT and text-only RL approaches.}
    \label{tab:appdx_baselines}
\end{table*}



\end{document}